\documentclass[twocolumn]{article}

\PassOptionsToPackage{dvipsnames,table}{xcolor}
\usepackage{preprint}
\SetBgContents{} 

\usepackage[utf8]{inputenc}
\usepackage[T1]{fontenc}
\usepackage[english]{babel}
\usepackage{csquotes}
\usepackage[final,
            tracking=true,
            kerning=true]{microtype}
\usepackage{xspace}
\usepackage{etoolbox}

\usepackage{amssymb}
\usepackage{amsmath}
\usepackage{amsthm}
\usepackage{mathtools}
\usepackage{bbm}

\usepackage{graphicx}
\usepackage{booktabs}
\AtBeginEnvironment{tabular}{\scriptsize}

\usepackage{nth}
\usepackage{siunitx}
\usepackage[backend=biber,
            sorting=nyt,
            maxcitenames=2,
            maxbibnames=8]{biblatex}
\usepackage{varioref}
\PassOptionsToPackage{hyphens}{url}
\usepackage[hypertexnames=false,
            colorlinks=true,
            linkcolor=gray,
            urlcolor=MidnightBlue,
            citecolor=gray,
            setpagesize=false]{hyperref}

\usepackage[all]{hypcap}
\usepackage[noabbrev,
            capitalize,
            nameinlink]{cleveref}

\let\cite\citep

\usepackage{xspace}

\newcommand*{\eg}{e.\,g.\@\xspace}
\newcommand*{\ie}{i.\,e.\@\xspace}

\newcommand*{\imgpix}{x}
\newcommand*{\cellcount}{l}
\newcommand*{\imgtype}{t}
\newcommand*{\typenat}{\mathfrak{n}}
\newcommand*{\typesynth}{\mathfrak{s}}
\DeclareMathOperator{\recocost}{Rec}
\DeclareMathOperator{\regrcost}{Reg}
\DeclareMathOperator{\regucost}{\mathcal{D}_{KL}}
\newcommand*{\weightparam}{C}
\DeclareMathOperator{\twincost}{Twin_{loss}}

\newcommand*{\dataset}[1]{\texttt{#1}}

\newcommand*{\dsnat}{\dataset{Nat}}
\newcommand*{\dssyn}{\dataset{Syn}}
\newcommand*{\dsbf}{\dataset{BF}}
\newcommand*{\dspc}{\dataset{PC}}
\newcommand*{\dstr}{\dataset{Tr}}
\newcommand*{\dste}{\dataset{Te}}
\newcommand*{\dsu}{\dataset{U}}
\newcommand*{\dsl}{\dataset{L}}

\newcommand*{\dsnatbf}{\dataset{Nat-BF}}
\newcommand*{\dsnatbfltr}{\dataset{Nat-BF-L-Tr}}
\newcommand*{\dsnatbfutr}{\dataset{Nat-BF-U-Tr}}
\newcommand*{\dsnatbflte}{\dataset{Nat-BF-L-Te}}
\newcommand*{\dsnatbfute}{\dataset{Nat-BF-U-Te}}

\newcommand*{\dsnatpc}{\dataset{Nat-PC}}
\newcommand*{\dsnatpcltr}{\dataset{Nat-PC-L-Tr}}
\newcommand*{\dsnatpcutr}{\dataset{Nat-PC-U-Tr}}
\newcommand*{\dsnatpclte}{\dataset{Nat-PC-L-Te}}
\newcommand*{\dsnatpcute}{\dataset{Nat-PC-U-Te}}

\newcommand*{\dssynbfltr}{\dataset{Syn-BF-L-Tr}}
\newcommand*{\dssynbflte}{\dataset{Syn-BF-L-Te}}

\newcommand*{\dssynpcltr}{\dataset{Syn-PC-L-Tr}}
\newcommand*{\dssynpclte}{\dataset{Syn-PC-L-Te}}

\newcommand*{\method}[1]{\texttt{#1}}

\newcommand*{\metcv}{\method{Watershed}}
\newcommand*{\meteff}{\method{EfficientNet}}
\newcommand*{\metebz}{\method{Efficient\-Net-B0}}
\newcommand*{\metebo}{\method{Efficient\-Net-B1}}
\newcommand*{\metvae}{\method{Twin-VAE}}
\newcommand*{\metvaemaxacc}{\method{Twin-VAE\textsubscript{max-acc}}}
\newcommand*{\metvaemindev}{\method{Twin-VAE\textsubscript{min-dev}}}

\newcommand*{\na}{n/a}
\newrobustcmd\best{\DeclareFontSeriesDefault[rm]{bf}{b}\bfseries}

\title{Towards an Automatic Analysis of CHO-K1 Suspension Growth in Microfluidic Single-cell Cultivation}

\usepackage[auth-sc]{authblk}

\author{Dominik Stallmann}
\author{Jan P. Göpfert}
\author{Julian Schmitz}
\author{Alexander Grünberger}
\author{Barbara Hammer}
\affil{Bielefeld University, Germany}

\begin{document}
\twocolumn[
\begin{@twocolumnfalse}
\maketitle
\begin{abstract}
\textbf{Motivation:} Innovative microfluidic systems carry the promise to greatly facilitate spatio-temporal analysis of single cells under well-defined environmental conditions, allowing novel insights into population heterogeneity and opening new opportunities for fundamental and applied biotechnology.
Microfluidics experiments, however, are accompanied by vast amounts of data, such as time series of microscopic images, for which manual evaluation is infeasible due to the sheer number of samples.
While classical image processing technologies do not lead to satisfactory results in this domain, modern deep learning technologies such as convolutional networks can be sufficiently versatile for diverse tasks, including automatic cell counting as well as the extraction of critical parameters, such as growth rate.
However, for successful training, current supervised deep learning requires label information, such as the number or positions of cells for each image in a series;
obtaining these annotations is very costly in this setting.\\
\textbf{Results:} We propose a novel machine learning architecture together with a specialized training procedure, which allows us to infuse a deep neural network with human-powered abstraction on the level of data, leading to a high-performing regression model that requires only a very small amount of labeled data.
Specifically, we train a generative model simultaneously on natural and synthetic data, so that it learns a shared representation, from which a target variable, such as the cell count, can be reliably estimated.\\
\textbf{Availability:} The project is cross-platform, open-source and free (MIT
licensed) software. We make the source code available at \url{https://github.com/dstallmann/cell_cultivation_analysis}; the data set is available at \url{https://pub.uni-bielefeld.de/record/2945513}\\
\textbf{Contact:} \href{mailto:dstallmann@techfak.uni-bielefeld.de}{dstallmann@techfak.uni-bielefeld.de}

\end{abstract}
\vspace{0.5cm}
\end{@twocolumnfalse}
]

\let\cite\autocite

\section{Introduction}

New and improved single-cell analysis technologies employing live cell imaging allow researchers to acquire cellular information with ever-increasing levels of detail~\cite{Grunberger2014}.
Especially for the study of cellular heterogeneity in clonal populations, which has long been ignored, single-cell analysis is mandatory.
Here, microfluidic single-cell cultivation~(MSCC) in particular is a promising tool to cultivate cells and analyze single-cell behavior~\cite{Schmitz2019}.
Due to its microfluidic setup, MSCC allows for maintaining clearly defined cultivation environments or even dynamic changes of cultivation conditions~\cite{Kolnik2012}.
Furthermore, MSCC encompasses live cell imaging and thereby allows systematic single-cell studies with high spatial and temporal resolution of cellular behavior.
Despite its beneficial features for single-cell analysis, MSCC comes with a major bottleneck:
enormous amounts of detailed data are generated that need processing.

So far, image analysis has mostly been performed manually or semi-manually.
Therefore, MSCC requires adaptations with regards to the processing of the data produced.
This either means the training of human experts and resulting extensive labor, or the development of new algorithms, where current approaches typically require either careful tuning or manually labeled data for training.
In the long run, this is not feasible, and different, more affordable computer vision solutions are required~\cite{Theorell2019}.

In recent years, deep neural networks, in particular convolutional neural networks, which naturally mirror spatial priors, have been established as the standard technology for automated
image processing and computer vision~\cite{Ioannidou2017}, and a considerable number of applications can be found in the biomedical domain~\cite{Razzak2018}.
While trained deep network models such as YOLO, VGG, or ResNet are readily available for natural scenes, these models do not easily transfer to other domains or imaging technologies, due to the different statistical properties observed in these fields.

There exist a number of approaches that tackle the challenge to track cells in images~\cite{Moen2019}:
Fundamental progress has been made by proposing a benchmark suite based on different imaging technologies and comparing the strengths and limitations of diverse methods for cell tracking~\cite{Ulman2017}.
Still, one conclusion is the observation that automatic tracking remains prone to errors, and the tuning of model parameters for novel domains can be very cumbersome.
The approach \cite{Berg2019} enables an  interactive bioimage analysis framework for intuitive user interaction. 

The work \cite{Hughes2018} focuses on the challenge of gathering labeled data by providing a crowd based annotation tool, which allows a distribution of the work of manual labeling, based on which deep learning becomes possible. Interestingly, the work \cite{Brent2018} displays the possibility to transfer results in between different imaging technologies to some extent. The approach \cite{Falk2019} provides one of the few toolboxes for cell tracking, which allows transfer learning based on given models and novel data, whereby data set enrichment technologies limit the number of required samples. Yet, the data used consists of adherent cells rather than suspension cells. 

In contrast to already reported single-cell cultivation studies \cite{Carlo2006} and \cite{Kolnik2012}, where adherent growing cell lines were the focus of investigation, in this study we will concentrate on suspension cells, for which analysis tools of adherent cells are insufficient: suspension cells have a circular basic shape but ever-changing contour due to vesicle secretion and the addition of cell movement and floating within the chamber complicates tracking and analysis processes.
Automatic analysis of these cells growing in suspension comes with different and challenging obstacles, which will be described in \cref{sec:data}.

Due to this fact, we cannot transfer existing models, such as the ones mentioned above, to the given setting.
We are interested in the question how to automatically provide sufficiently accurate cell counting, based on which the process dynamics can be characterized.
We will use deep convolutional networks for this task.
Unlike existing technologies, which are typically fully supervised, we put a particular focus on the challenge to mitigate the cumbersome task of manual labeling.

We propose a new deep twin auto-encoder architecture called \metvae{}, which enriches the training data by artificial geometric data, for which ground truth labeled data can easily be generated.
The proposed twin architecture greatly reduces the cost of synthetic, auxiliary training data, because that data does not need to appear realistic in all regards, such as morphological details and fidelity.

Training is based on a novel cost function that enriches the final task of cell counting by the challenge of correct representations.
Optimization of this architecture is partially based on auto-ML technologies~\cite{He2020}.

\section{Material and Methods}
\label{sec:data}
\subsection{MSCC and live cell imaging data}

\begin{figure}[t!]
\centering
\includegraphics[width=\linewidth]{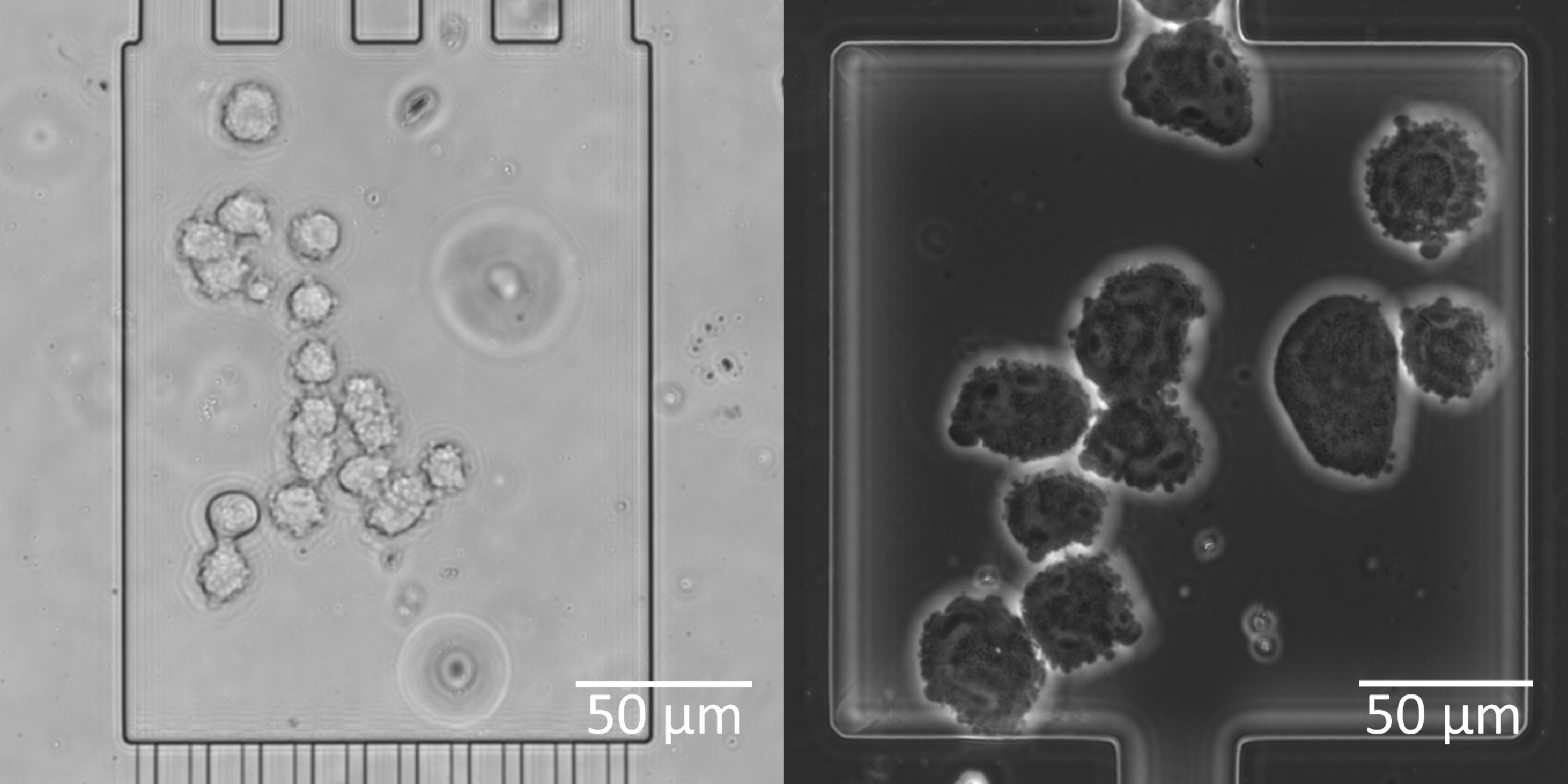}
\caption{\ Samples from the data sets. Bright-field microscopy image on the left, phase-contrast microscopy image on the right. Faint and very small circles are smudges on the chip.
The architecture has to differentiate between cells, smudges and background.
Translational and rotational camera drift has been removed by semi-automatic image processing beforehand.
The crop intentionally allows for data augmentation techniques to crop the images further. The scale bar has been added manually and is not part of the data set.}\label{fig:sample}
\end{figure}

Image data applied in this study was obtained by single-cell cultivation of mammalian suspension cells as shown before ~\cite{Schmitz2020}.
CHO-K1 cells were cultivated in polydimethylsiloxane (PDMS)-glass-chips and constantly provided with nutrients by perfusion of the microfluidic device.
Live cell imaging was accomplished with an automated inverted microscope (Nikon Eclipse Ti2, Nikon Instruments, Germany), every \num{20}~minutes time lapse images of relevant positions on-chip were taken (NIS Elements AR 5.20.01 Software, Nikon Instruments, Germany).
The data used in this work is split into two major parts according to the two microscopy technologies, namely bright-field microscopy and phase-contrast microscopy, that were used for the analysis of the cells (see \cref{fig:sample}). 

\cref{Tab:01} shows an overview of all data sets. The sets declared as \texttt{Nat} contain the natural images mentioned above. Of these, \texttt{BF} declares bright-field microscopy images which consist of \num{12}~scenarios accumulating to \num{956}~used images, of which \SI{7.5}{\percent} of the training images are labeled, while \texttt{PC} denotes phase-contrast microscopy images consisting of \num{37}~scenarios and accumulates to \num{3976}~used images of which \SI{6.2}{\percent} of the training images are labeled.
The labels are created in a regular interval over the data sets and experiment scenarios.
Images with more than \num{30}~cells were removed beforehand, since at this cell count, the following trend of the cultivation experiment is determined.

\begin{figure}[t]
\centering
\includegraphics[width=\linewidth]{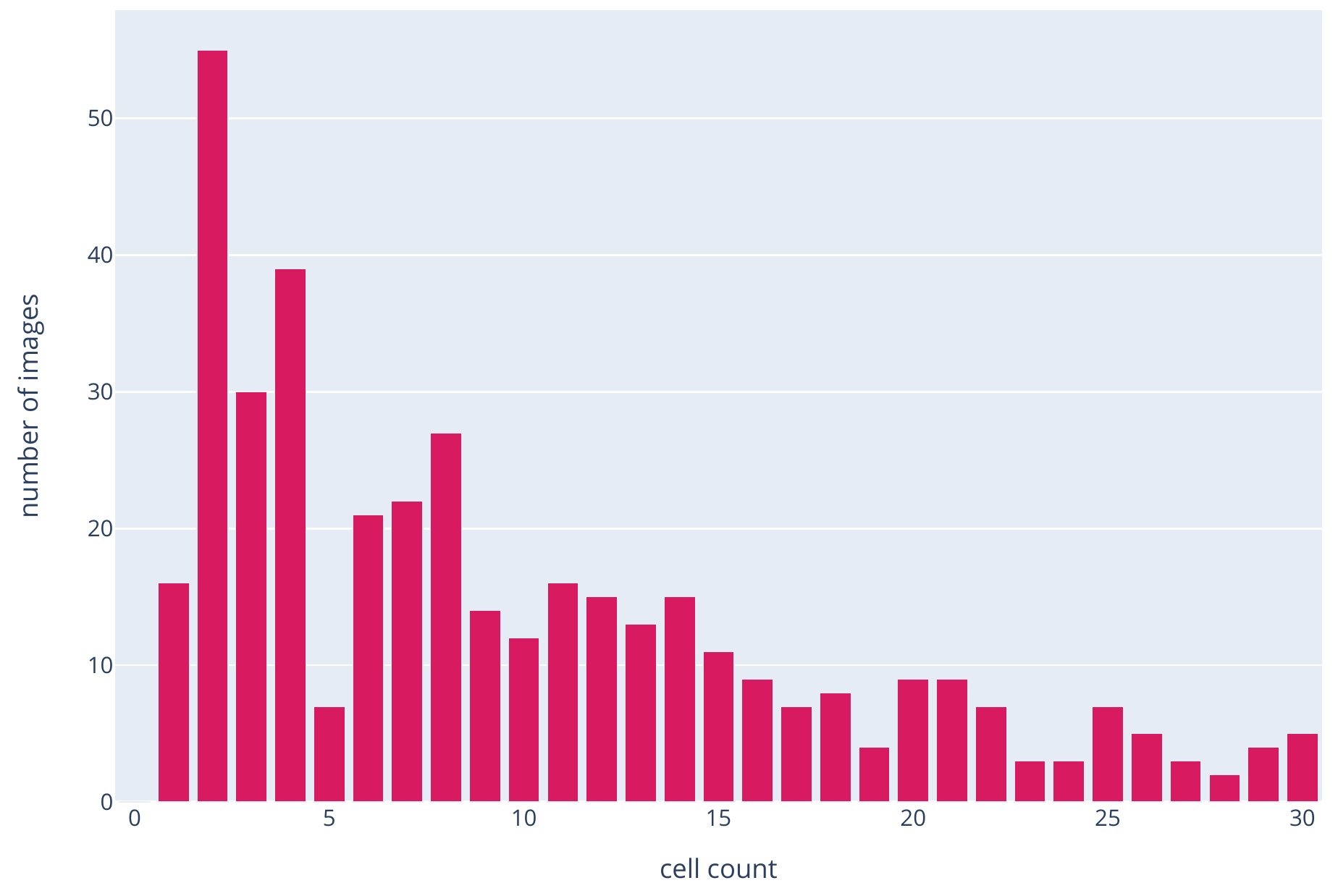}
\caption{\ Distribution of images by cell count for \dsnatpclte{} and \dsnatpcute{}. We omit images that contain more than \num{30}~cells because those are irrelevant for our cultivation experiments.}\label{fig:distribution}
\end{figure}

We put our focus on the larger data set \dsnatpc{}.
Not only does it contain more samples,
but the captured biological processes are more varied,
and phase-contrast microscopy is arguably more popular than bright-field microscopy.
We show the distribution of images against the cell counts in them in \cref{fig:distribution} for \dsnatpc{}.
For both \dsnatbf{} and \dsnatpc{},
we have labeled a number of images by hand, but we use most labels to test (\dataset{L-Te}) the reconstruction, rather than during training (\dataset{L-Tr}),
because our goal is a method that does not rely on large amounts of labeled data.
We use further unlabeled test data (\dataset{U-te}) to evaluate the reconstruction.
We crop and rotate all images to center the cultivation chamber.
Further data augmentation beyond this preprocessing is described in \cref{subsec:da}.

Among the more challenging aspects in the cultivation of suspension cells, in comparison to adherent cells, are the large number of visual characteristics, which prevent traditional cell counting techniques and tracking approaches~\cite{Riordon2019}.
As can be seen in \cref{fig:sample}, there are a multitude of visual effects that complicate the automated counting process.
Often, certain of these characteristics are being tackled, such as varying contrast and light conditions~\cite{Chen2017}, or cells sticking together and overlapping each other~\cite{Weidi2018}, but additional ongoing cell divisions and changes in shape due to secretion of vesicles, which makes the cells' silhouettes highly irregular, impede the process and diminish the applicability of state-of-the-art solutions.
Furthermore, appearing and vanishing of cells through the entrance of the chamber, a potential overpopulation of the culture and differing focus plains based on the cells’ dimensions and inner organelles, make intracellular compartment non-uniform, further increasing the challenge to automate counting.

\begin{table}[!b]
\caption{Overview of our data sets. \dataset{Nat} and \dataset{Syn} indicate natural and synthetic data. \dataset{BF} and \dataset{PC} indicate bright-field and phase-contrast. \dataset{L} and \dataset{U} indicate labeled and unlabeled data. \dataset{Tr} and \dataset{Te} indicate training and test data.}
\label{Tab:01}
{
\centering
\begin{tabular}{@{}cccccS[table-format=4.0]@{}}\toprule
Abbreviation & Type      & Technique      & Labeled & Use      & {Size} \\\midrule
\dsnatbfltr  & natural   & bright-field   & yes     & Training &     59 \\
\dsnatbfutr  & natural   & bright-field   & no      & Training &    725 \\
\dsnatbflte  & natural   & bright-field   & yes     & Testing  &     62 \\
\dsnatbfute  & natural   & bright-field   & no      & Testing  &    110 \\
\dsnatpcltr  & natural   & phase-contrast & yes     & Training &    199 \\
\dsnatpcutr  & natural   & phase-contrast & no      & Training &   2983 \\
\dsnatpclte  & natural   & phase-contrast & yes     & Testing  &    397 \\
\dsnatpcute  & natural   & phase-contrast & no      & Testing  &    397 \\
\dssynbfltr  & synthetic & bright-field   & yes     & Training &   1000 \\
\dssynbflte  & synthetic & bright-field   & yes     & Testing  &   1000 \\
\dssynpcltr  & synthetic & phase-contrast & yes     & Training &   3182 \\
\dssynpclte  & synthetic & phase-contrast & yes     & Testing  &    794 \\\bottomrule
\end{tabular}
}
\end{table}

\subsection{Approach}
Our goal is reliable cell counting for suspension cell microscopic images.
Since given data is limited and with only few manual annotations,
we construct a novel deep neural network architecture (described in \cref{sec:model}) to overcome these limitations.
The challenge to learn from mostly unlabeled data, which are enriched by few manual labels, is usually referred to as semi-supervised learning \cite{CGoepfert2019}.
We will propose a novel technology to mitigate this complexity, by introducing synthetic data with known ground truth and a twin architecture, which can abstract from the fact that these enrichments are synthetic.
We will now describe how we generate the data set enrichment, first, and introduce the specific neural architecture thereafter.

Since semi-supervised learning is a balancing act between reducing the share of labels required for acceptable results and increasing the share of labels to achieve desirable results~\cite{Deepak2019}, an additional dimension can be opened to tackle the problem from another side.
We propose to generate an auxiliary training set \dssyn{} with simplified synthetic data, for which the ground truth is known because it is based on explicit geometric modeling, thereby simplifying the geometric heterogeneity of this data compared to real suspension cells drastically.
Synthetic data has already successfully been used to improve the performance of artificial neural networks, \eg, for text processing~\cite{Jaderberg2014}. 
Unlike data set enrichment, which directly enriches given data, synthetic data enables us to create a large variety of samples, which is not limited by the real data.

One might try to synthetically generate images that are indistinguishable from real ones, which would entail a considerate amount of engineering~\cite{Gopfert2017VariabilitySynthetic}.
Instead, we rely only on basic geometric shapes to represent cells, neglecting concrete texture and intricate morphology.
In the following, we show that this is adequate for successful training using our \metvae{} architecture (\cref{sec:model}) and it actually allows us to introduce our abstraction into the data:
cells (in these kinds of images) can be represented by ellipses, especially when the objective is to count them.

\begin{figure}[t!]
\centering
\includegraphics[width=\linewidth]{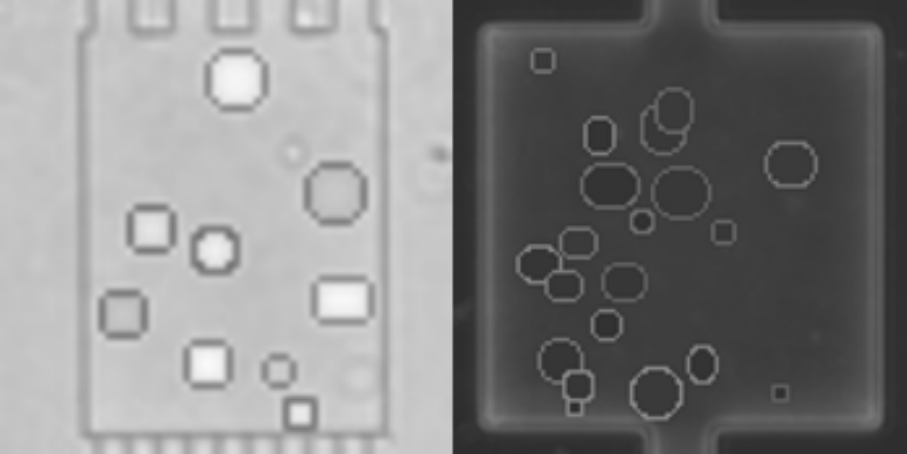}
\caption{\ Samples of synthetic data from the proxy data sets \dssynbfltr\  and \dssynpcltr.
Bright-field microscopy image on the left, phase-contrast microscopy image on the right.
The images do not show scale bars, because they are dimensionless.
A theoretical size could be calculated in relation to the chamber size, but it is of no importance for the work.}\label{fig:sample_syn}
\end{figure}

Examples of the synthetic images can be seen in \cref{fig:sample_syn}. \cref{Tab:01} lists this data as \dssyn{}, followed by the microscopy technology categorization \dsbf{} or \dspc{} respectively, omitting the \dsu{} or \dsl{} declaration since all synthetically generated data is labeled and differentiates between independently generated training (\dstr{}) and test (\dste{}) images.
The cell distribution in these images is close, but not identical, to that of the \dsnat{} data sets.
These images are generated algorithmically with seed consistency, are thus reproducible and can be generated in arbitrary amount, however larger amounts of synthetic data will increase training time nearly linearly, while being assumed to improve performance with diminishing returns. For that and the sake of balance, the \dssyn{} set sizes are determined to roughly match the number of natural images.
The background is created by taking the mean of all natural training images with few cells (to ensure high visibility of the background).
The synthetic images are generated in the \num{128}~by \num{128}~pixels working resolution of the architecture described in \cref{sec:model} and their size is about \SI{8}{MB} per \num{1000}~images.

The generator is fully adjustable, creating images with a given distribution of cell counts, controls overlapping of cells, varies the brightness of the cell's inner organelles their membrane silhouette and the background, etc., such as in the natural data. Combined, these controls can also visualize the more complex visual aspects of natural data, such as ongoing cell divisions, by creating a small overlap together with more noisy cell borders. Smudges, as in \cref{fig:sample}, have not been inserted, since they are an interference factor and are assumed to only hinder the training process.
The cells have been given a circular shape to roughly match the shape of the natural cells.
Some stretching or deforming to ellipses, noise, brightness fluctuation and Gaussian filters of varying strengths have been added to increase the variety of cells in the data.
This geometric form can easily be adjusted if natural cells in other data sets have different shape characteristics.

\subsection{Network Architecture \& Training}
\label{sec:model}
\subsubsection{Architecture}
\begin{figure*}[t]
\centering
\includegraphics[width=1\linewidth]{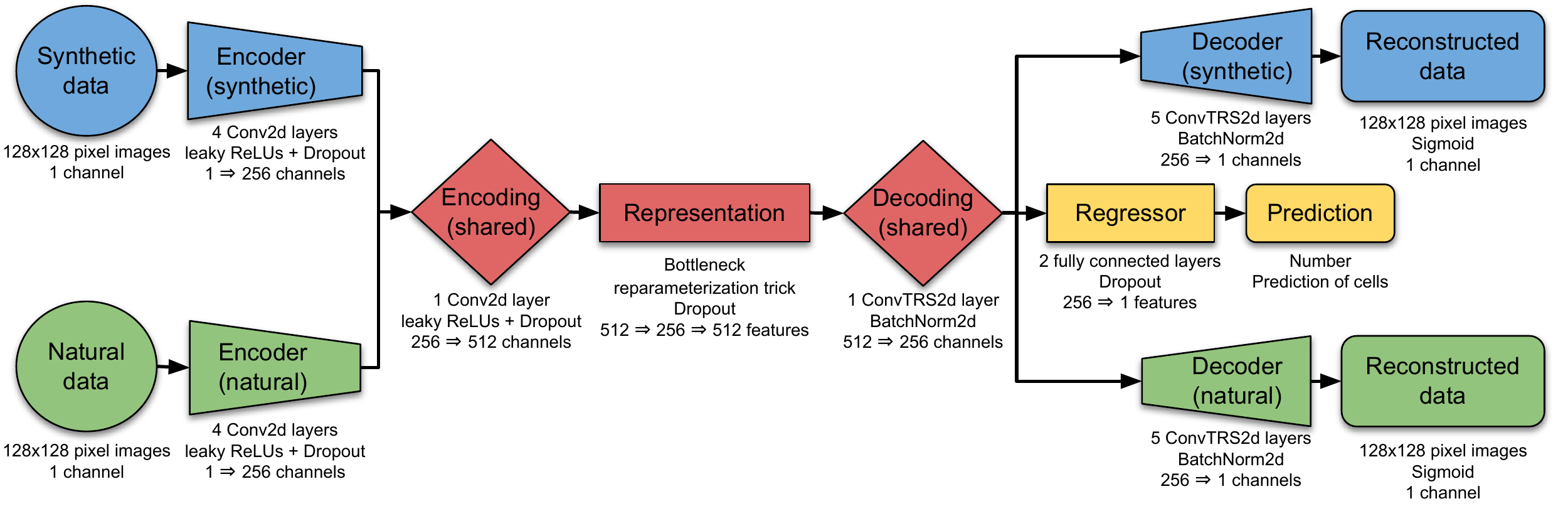}
\caption{Visualization of the \metvae{} architecture.
The blue elements handle synthetic data, while the green elements handle natural data.
The red elements are shared between the two VAEs and contain the inner representation of the cell imagery, while the yellow elements result in an estimation for the cell count.}\label{fig:architecture}
\end{figure*}

We propose a novel architecture that bypasses the problem of difference in appearance of synthetic and natural images, by separating the input of data for training depending on their origin, but requiring the architecture to create a tightly coupled shared inner representation to prevent high training losses.
We do so by creating two identical variational autoencoders (VAE) for the two data sets, sharing the weights of their last layer of the encoder, the first layer of the decoder and the bottleneck in-between (see \cref{fig:architecture}).
VAEs are the state-of-the-art solution for generalized few-shot learning~\cite{Schonfeld2019} and weight-sharing has been used to reduce neural network sizes and to improve test performance before~\cite{Karen2017}.

One of the VAEs uses synthetic data (\textit{VAE-syn}), while the other one handles natural data only (\textit{VAE-nat}).
The different visual characteristics of synthetic and natural data are accounted for in the non-shared layers, while the shared layers rely on and enforce a common representation of relevant image characteristics.
Besides auto-encoding, the architecture acts in a supervised way for data with known labels, by
adding a two-layer-deep fully connected neural network regression model for the actual cell counting, based on the shared representation of the VAEs.
Cell detection by regression has been shown to work well for other (less demanding) tasks~\cite{Yuanpu2015,Weidi2018}.

Our architecture addresses two objectives simultaneously:
\begin{itemize}
  \item Encoding as well as decoding natural and synthetic input images using a shared representation. 
  \item Predicting the present cell count for both types of images.
\end{itemize}
\subsubsection{Loss}
Given an input image \(\imgpix\) of pixels, a label (\ie, cell count) \(\cellcount\) between \num{1} and \num{30} and a type \(\imgtype \in \{\typenat, \typesynth\}\), representing the fact whether the image is natural or synthetic, we obtain a reconstruction loss \(\recocost(\imgpix)\) of the VAE, a regression loss \(\regrcost(\imgpix, \cellcount)\) of the task at hand such as cell counting, and a distributional regularization loss \(\regucost\), which aims for a homogeneous representation of synthetic and real data in the embedding space of the VAE.
We combine these losses to form our twin loss \(\twincost(\imgpix, \cellcount, \imgtype)\) with weighting factors \(\weightparam_{\recocost}^{\imgtype}\), \(\weightparam_{\regrcost}^{\imgtype, \cellcount}\), and \(\weightparam_{\regucost}^{\imgtype}\), respectively, which allows us to balance the impacts between natural and synthetic images and to gracefully handle input images without known cell counts by setting \(\weightparam_{\regrcost}^{\imgtype, \cellcount}\) to zero:
\begin{equation}\label{eq:twinloss}
\begin{split}
    \twincost(\imgpix, \cellcount, \imgtype) = \\
    \weightparam_{\recocost}^{\imgtype} \cdot \recocost(\imgpix)
    + \weightparam_{\regrcost}^{\imgtype, \cellcount} \cdot \regrcost(\imgpix, \cellcount)
    + \weightparam_{\regucost}^{\imgtype} \cdot \regucost(\imgpix)
\end{split}
\end{equation}

During our experiments, the mean-squared error (MSE) \(||x-d(x)||^2\), where \(d(x)\) is the reconstruction of the input image \(x\) and \(||l-r(x)||^2\), where \(r(x)\) is the estimated cell count, yielded the best results respectively, when used as \(\recocost(\imgpix)\) and \(\regrcost(\imgpix, \cellcount)\) for training on phase-contrast data, and as \(\regrcost(\imgpix, \cellcount)\) for bright-field data. However, for bright-field data the binary cross entropy (BCE) \(-l\cdot log(r(x))+(1-l)\cdot log(1-r(x))\)
turned out to be the superior choice for \(\recocost(\imgpix)\) and was often resulting in just slightly worse results than the MSE for phase-contrast data.
The \(\regucost\) is applied in the form of the Kullback-Leibler divergence (KLD) of the standard VAE (\cite{Kingma2013}) and is needed to generate latent vectors that are sufficiently close to a normal distribution.
The weighting factors \(\weightparam_{\recocost}^{\imgtype}\), \(\weightparam_{\regrcost}^{\imgtype, \cellcount}\), and \(\weightparam_{\regucost}^{\imgtype}\) have to be chosen carefully for training to succeed, punishing incorrect cell count predictions especially on natural cells, while relaxing the importance of visual reconstruction.
We provide details in the following section.

\subsubsection{Neural Network}
\label{subsec:nn}
The non-shared part of the encoder consists of four two-dimensional convolutional layers with kernel size \num{5} and stride \num{2}, and are initialized with an orthogonal basis~\cite{Saxe2014}.
Between the layers are leaky rectified linear units (ReLU) with a leakiness of \num{0.2} as well as dropout of \num{0.1}.
The amount of channels used for the convolutions are \num{32}, \num{64}, \num{128} and \num{256}, in this order, for the encoders.
The shared part of the encoder has an additional two-dimensional convolutional layer with the same properties and \num{512} channels.
The layer is followed by the bottleneck, which consists of three layers of fully connected neurons of sizes \num{512}, \num{256} and \num{512} again, each with a dropout of \num{0.1}.
The shared part of the decoder has \num{256}~channels, uses a two-dimensional transposed convolutional operator layer with the same kernel size and stride as above, and is followed by a batch normalization over a four-dimensional input and another leaky ReLU with a leakiness of \num{0.2}.
The non-shared part of the decoder consists of five layers of kernel sizes \num{5}, \num{5}, \num{5}, \num{2}, \num{6}, a continuous stride of \num{2} except for the fourth layer with a stride of \num{1}, the same leaky ReLUs and a sigmoidal activation function at the end.
The output of the shared part of the decoder is also fed to a two-layer fully connected branch of neurons for the regressor of sizes \num{256} and \num{128}.
The regressor uses linear layers and a dropout of \num{0.1}.

The architecture is using the Adam optimizer for phase-contrast microscopy data, and the rectified Adam (RAdam)~\cite{Radam2020} optimizer for bright-field data.
The combination of the decoder loss factor \(\weightparam_{\recocost} = 100\), the regressor loss factor \(\weightparam_{\regrcost}=3\) and the KLD factor \(\weightparam_{\regucost}=2\) yield the best results for phase-contrast data.
For the BCE, the decoder loss factor is not constant, but decays over time with a rate of \num{3e-5} per epoch, since the BCE will not decrease significantly within the training process, but needs to decrease over time to amplify the importance of low regression losses \(\regrcost(\imgpix, \cellcount)\).

While it seems counter-intuitive that \(\weightparam_{\recocost}\) is bigger than \(\weightparam_{\regrcost}\) and \(\weightparam_{\regucost}\), it is caused by the MSE for pixel data getting very small on normalized images and it's actually being a desirable factor for the KLD to stay relatively small, since it is required to enhance the quality of the distributions, but should not impact the training of cell predictions too much by unfortunate sampling from the latent vector.
Additionally a soft weight decay of \num{1e-5} per epoch, a fixed learning rate of \num{1.3e-4} and a delayed start for the regressor of \num{100}~epochs were used to achieve the results presented in \cref{sec:evaluation}.
A batch size of \num{128} for the phase-contrast images and \num{64} for the bright-field images worked best and the training runs for up to \num{50.000} epochs, unless early stopping conditions abort it.

To minimize the number of hyperparameters that had to be optimized by hand, a Bayesian optimization~\cite{Bayes2011} in form of a Gaussian Process regressor~\cite{Williams1996} has been used to find good values for the learning rate, the number of channels of the bottleneck and the convolutional layers around it, as well as all three loss weight factors.

\subsubsection{Data Augmentation}
\label{subsec:da}
To maximize the use of the limited amounts of natural data, multiple data augmentation techniques were combined and applied to the data.
Random horizontal and vertical flips are combined with a random resized crop of scale ~\num{0.9}, meaning the images get randomly cropped to \num{115} by \num{115} pixels and then scaled back to \num{128} by \num{128} pixels.
While the crop does add difficulty to the cell detection process by allowing cells to be at positions where, without the crop, only the chamber border and outside of the chamber would be, it proved necessary to establish the possibility of cells appearing anywhere on the image, assuring equalized detection success, barely impacted by the cell position within the image.
Then, a \num{90} degree rotation is applied at random and a zero-centered noise map is generated and added to the image with a small amplitude factor.

\subsection{Baselines}
\label{sec:baselines}
We implement two different methods to serve as baselines for our evaluation.

The first is a state of the art classical computer vision pipeline.
First, the input images is cropped to only contain the cell chamber and blurred with a averaging kernel-based filter, then a thresholding filter is applied, followed by a watershed segmentation~\cite{Watershed2013}. The regions of the segmented image are counted and used as cell estimation.
To find good parameters for this pipeline, we performed an exhaustive grid search for each of the two data sets \dsbf{} and \dspc{}. Our code repository contains the optimal parameters found.
Because watershedding is the core component of this pipeline, we henceforth refer to it as \metcv{}.

As our second baseline, we fine-tune a pretrained state-of-the-art deep convolution neural network, specifically a variant of the \emph{EfficientNet}~\cite{Tan2019}.
We replace the last layer of the pretrained network by a fully-connected layer that outputs a single value and train this to predict the cell count for a given input image.
We apply the same hyperparameter optimization as for our own method as well as the same data augmentation.
Again, our code repository contains the exact implementation.
\emph{EfficientNet} is a variable architecture that comes in different sizes,
referred to as \emph{EfficientNet-B0}, \emph{EfficientNet-B1}, and so forth.
We tested \emph{EfficientNet-B0} through \emph{EfficientNet-B3} and found the smallest variant \emph{EfficientNet-B0} to perform best,
with larger variants performing progressively worse.
We henceforth refer to this fine-tuned convolutional neural network as \meteff{} with a suffix to indicate the respective variant.

\section{Results}
\label{sec:evaluation}
\subsection{Cell Counting}
We present the results of all methods on the data sets \dssynpclte{}, \dsnatpclte{}, \dssynbflte{}, and \dsnatbflte{} in \cref{table:results-cell-counts}.
Our \metvae{} consistently outperforms the other methods \metcv{} and \meteff{} by a wide margin.
Even without any synthetic data, our architecture leads to respectable results (\metvae{} (\dsnat{} only));
however, only through the simultaneous training on synthetic data do we realize its full potential.

We found different variants of our \metvae{} to out-perform each other depending on whether we optimize for low deviation in cell counts or high accuracy (meaning correct cell counts) on natural data (\dsnat{})
Therefore, we include two variants in \cref{table:results-cell-counts},
namely \metvaemaxacc{} optimized for high accuracy,
and \metvaemindev{} simultaneously optimized for low mean absolute error (MAE) and low mean relative error (MRE).
The MRE normalizes the error by the true cell count,
so that for high cell counts,
small deviations are not punished as much as they are for low cell counts.

\metvae{} estimates cell counts correctly for around \SI{58}{\percent} of all images in \dsnatpclte{} and it's count is on average only \num{0.6}~cells off the actual cell count in the images (MAE).
and it achieves approximately \SI{0.06}{\percent} mean relative error (MRE).
As such, \metvae{} proves capable of reliably counting cells in different microscopy technologies.

\begin{figure}
\centering
\includegraphics[width=\linewidth]{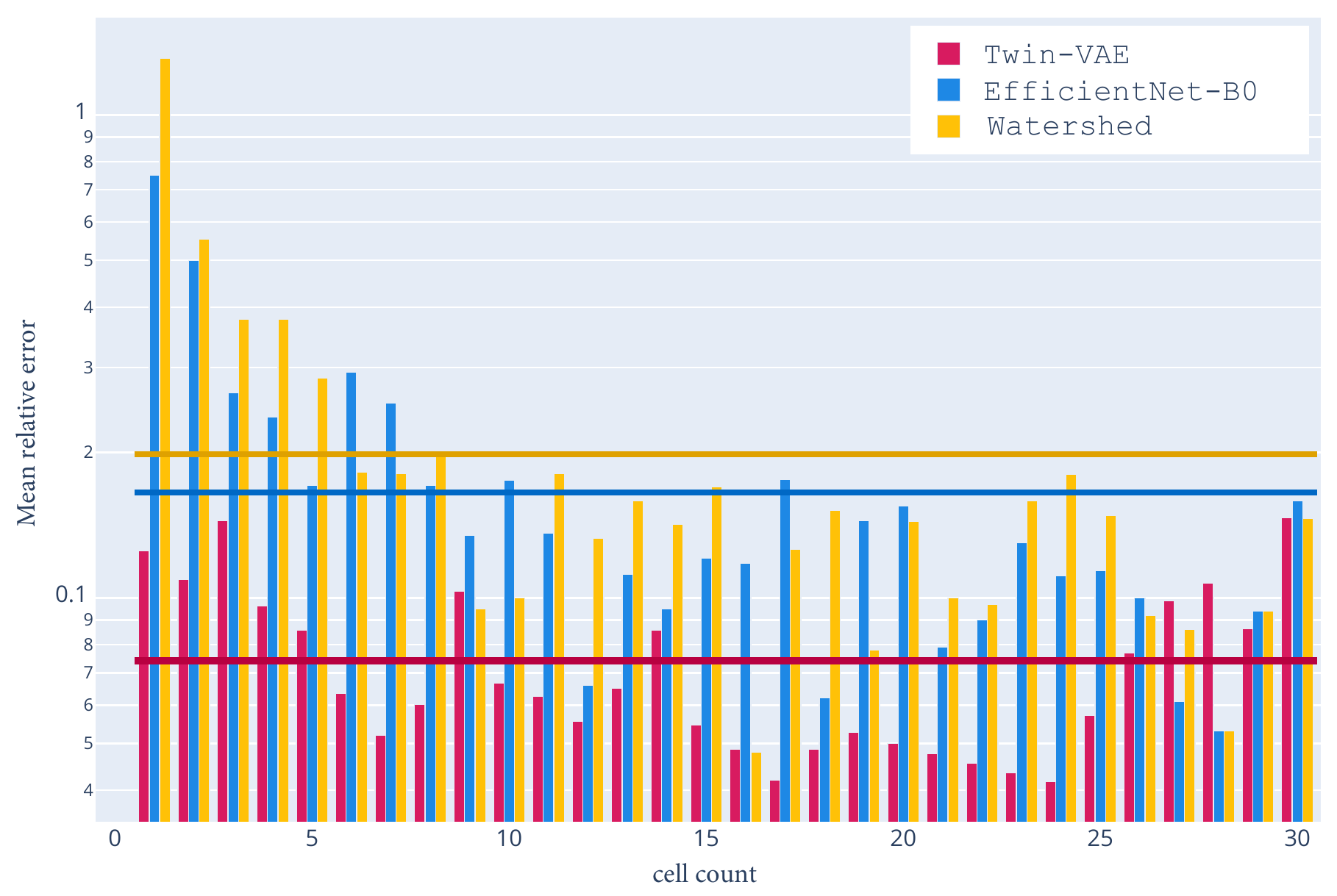}
\caption{\ The mean relative error (MRE) for \metcv{}, \metebz{}, and \metvaemindev{} on \dsnatbflte{} on a logarithmic scale. Horizontal bars indicate the average MRE of their respective color and method.}
\label{fig:deviation}
\end{figure}

We visualize the mean relative error for \metcv{}, \metebz{}, and \metvaemindev{} on \dsnatpclte{} in \cref{fig:deviation}.
The plot confirms the result from \cref{table:results-cell-counts}, showing the superiority our \metvae{} over the other methods.
Moreover, we see that \metvae{} performs well across the entire range of cell counts in \dsnatpclte{}.
By contrast, \metcv{} and \metebz{} struggle with images that contain few cells, which is the most important range of cell counts for the biological tasks, like estimating the growth rate.

\begin{table*}[!t]
\caption{Evaluation of all methods on the data sets \dssynpclte{}, \dsnatpclte{}, \dssynbflte{}, and \dsnatbflte{}.
For each method and data set, we report the mean absolute error (MAE), the mean relative error (MRE), and the accuracy.
Ultimately, only the performance on natural data (\dsnat{}) is important,
but we also report the performance on synthetic data (\dssyn{}) to provide further context.
We use an upward arrow \(\uparrow\) to indicate that higher is better
and a downward arrow \(\downarrow\) to indicate that lower is better.}
\label{table:results-cell-counts}
\centering
\sisetup{detect-weight,table-auto-round}
\begin{tabular}{cS[table-format=2.2]S[table-format=2.2]S[table-format=2.2]S[table-format=2.2]S[table-format=2.2]S[table-format=2.2]}\toprule
Method & {MAE (\dssyn{}) $\downarrow$} & {MRE / \% (\dssyn{}) $\downarrow$} & {Acc. / \% (\dssyn{}) $\uparrow$} & {MAE (\dsnat{}) $\downarrow$} & {MRE / \% (\dsnat{}) $\downarrow$} & {Acc. / \% (\dsnat{}) $\uparrow$}\\\midrule
\multicolumn{7}{c}{{\dspc{} (phase-contrast microscopy)}} \\
\metcv{}                  &       0.94 &       18.0 &       24.0 &       1.66 &       29.0 &       23.1 \\
\metebz{}                 &      4.987 &       79.4 &        5.0 &       1.67 &       25.1 &       23.4 \\
\metebo{}                 &      3.525 &       54.5 &        8.7 &       1.78 &       31.9 &       19.3 \\
\metvae{} (\dsnat{} only) &        \na &        \na &        \na &       1.07 &       20.1 &       39.8 \\
\metvaemaxacc{}           & \best 0.09 & \best 0.68 & \best 68.2 &       0.60 &       5.92 & \best 57.8 \\
\metvaemindev{}           &       0.14 &       0.73 &       62.1 & \best 0.59 & \best 5.66 &       57.0 \\
\midrule
\multicolumn{7}{c}{{\dsbf{} (bright-field microscopy)}} \\
\metcv{}                  &       1.92 &       39.0 &        2.0 &       2.39 &       32.0 &       32.0 \\
\metebz{}                 &      6.502 &       67.1 &        4.5 &       1.13 &       17.2 &       33.9 \\
\metebo{}                 &      5.245 &       67.5 &        4.4 &       1.21 &       18.5 &       29.0 \\
\metvae{} (\dsnat{} only) &        \na &        \na &        \na &       0.91 &       13.3 &       23.4 \\
\metvaemaxacc{}           & \best 0.48 & \best 4.27 & \best 60.1 &       0.68 &       7.60 & \best 53.2 \\
\metvaemindev{}           &       0.52 &       4.63 &       58.2 & \best 0.63 & \best 7.31 &       51.9 \\\bottomrule
\end{tabular}
\end{table*}

\subsection{Image Reconstruction}
While our goal is automatic counting of cells,
our loss \cref{eq:twinloss} includes a term for image reconstruction.
An analysis of the reconstruction abilities of \metvae{} is useful to ensure the learned shared representation is meaningful.

During training of \metvae{},
natural inputs are first processed by a specialized encoder,
then by a shared encoder and decoder,
and finally reconstructed by a specialized decoder (see \cref{fig:architecture}).
The equivalent is true for synthetic images.
For the regression to work as intended and for the cell counting in natural images to benefit from synthetic data,
the learned representation must be shared between the two types of data.
We can verify this by encoding a natural image with the appropriate encoder but performing the reconstruction with the decoder intended and trained for synthetic images -- or vice versa.
In the following, we demonstrate exactly this.

\begin{figure}
\centering
\includegraphics[width=\linewidth]{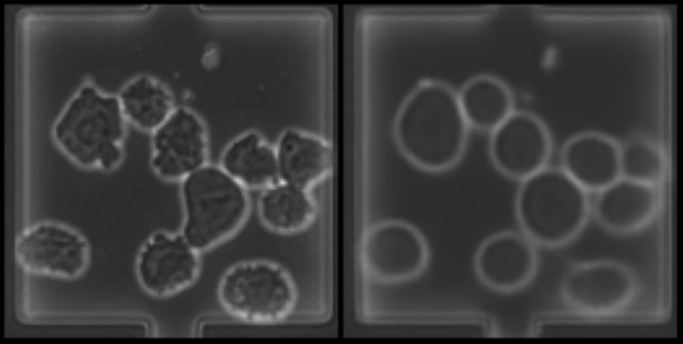}
\caption{Example of a perfect synthetic-looking reconstruction (right) of a natural image (left) from \dsnatpclte.
The cell counts match exactly and the position as well as size of cells are preserved.
While the top-right positioned smudge is recreated visually, it does not lead the regressional part of the \metvae{} to a wrong cell count.}
\label{fig:optimal_pc}
\end{figure}

\begin{figure}
\centering
\includegraphics[width=\linewidth]{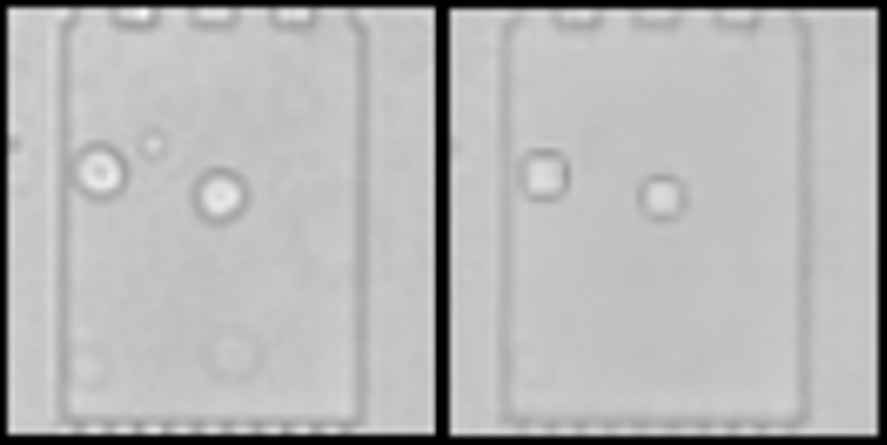}
\caption{Example of a perfect synthetic-looking reconstruction (right) of a natural image (left) from \dsnatbflte.
The cell counts match exactly and the position as well as size of cells are preserved. For this data set, where smudges are more faint, they don't get reconstructed usually.}
\label{fig:optimal_bf}
\end{figure}

In \cref{fig:optimal_pc} and \cref{fig:optimal_bf} we show examples of perfect translations,
where a natural image is encoded and subsequently decoded as a synthetic image.
The cell count is unchanged and the position and size of cells are also retained.
The overall appearance is simplified, though:
\metvae{} has learned to remove noise and condense the information required to count cells.


\begin{figure}
\centering
\includegraphics[width=\linewidth]{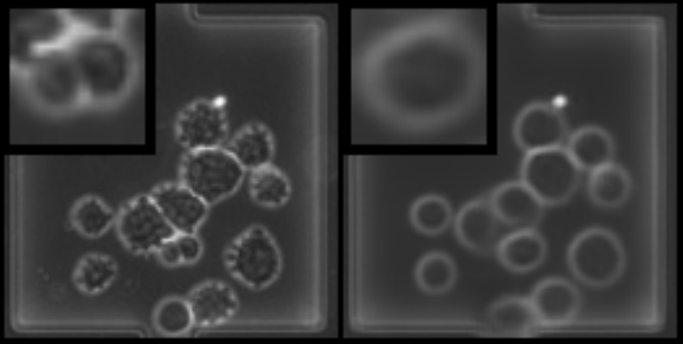}
\caption{Example of a faulty synthetic-looking reconstruction (right) of a natural image (left) from \dsnatpclte.
The cell count is off by one,
and we can visually spot the reason why:
two cells that are very close in the natural original are merged into a single cell.}
\label{fig:suboptimal}
\end{figure}

Even when \metvae{} does not translate an image perfectly,
the reconstruction can be useful to understand where an error occurs.
In \cref{fig:suboptimal} we show an example where two cells that are very close together are interpreted and reconstructed as a single cell.

\begin{figure}
\centering
\includegraphics[width=\linewidth]{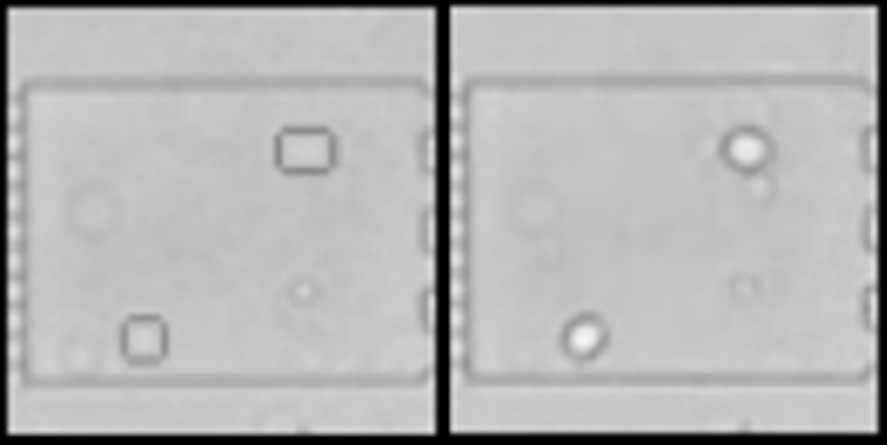}
\caption{Example of a perfect natural-looking reconstruction (right) of a synthetic image (left) from \dssynbfltr. These conversions ensure representational consistency on a visually comprehensible level. More details on this in \cref{fig:umap}.
The cell counts match exactly and the position and size of cells are preserved.}
\label{fig:syntonat}
\end{figure}

As well as translating images from natural to synthetic-looking,
\metvae{} can perform the inverse translation from synthetic to natural-looking as well.
We provide an example in \cref{fig:syntonat}. 

\subsection{Visualizing the Shared Representation}
\begin{figure}
    \centering
    \includegraphics[width=\linewidth]{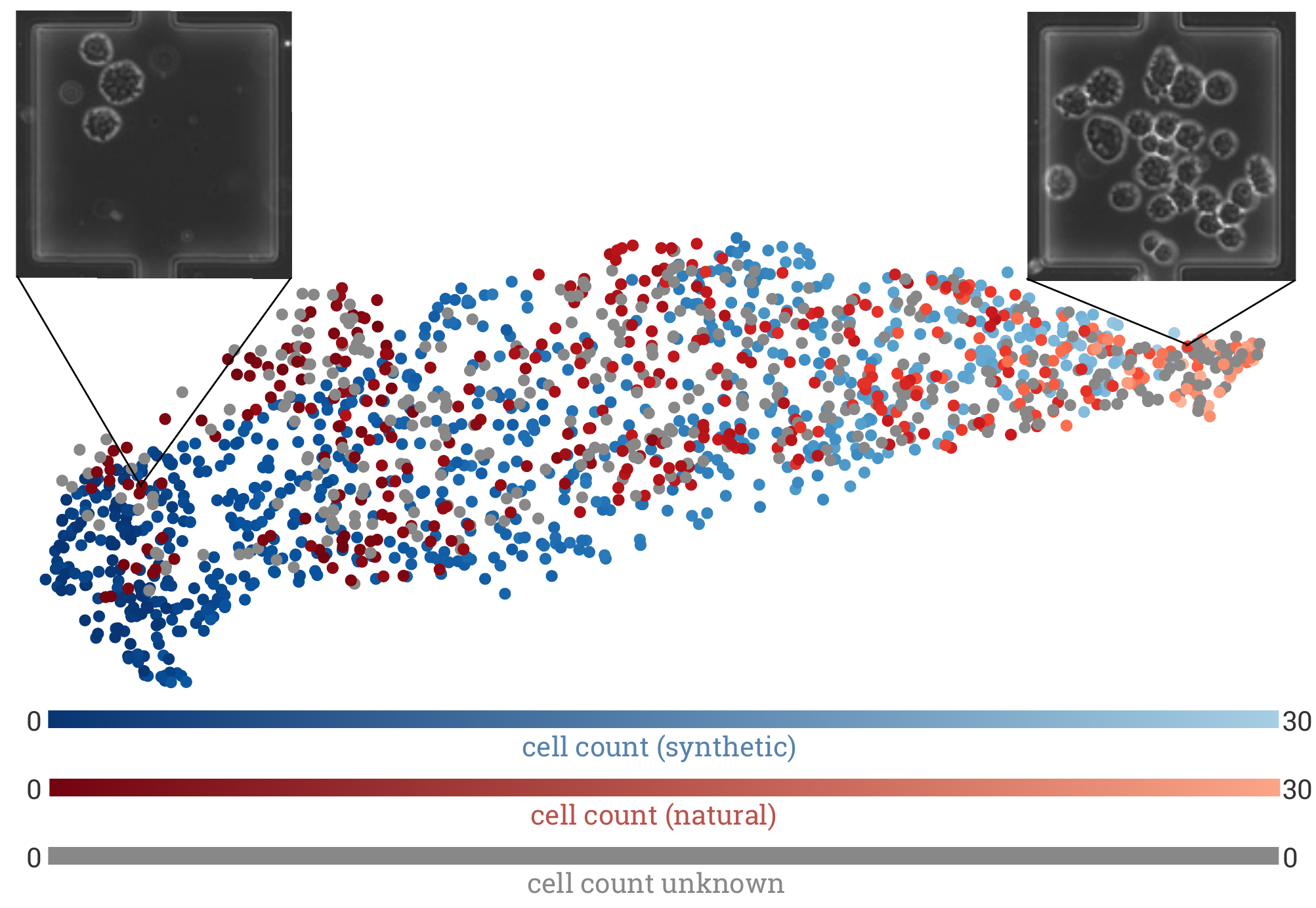}
    \caption{Embedding of learned representations, computed via UMAP.
    Visible are \num{392}~natural samples (red circles) and \num{794}~synthetic samples (blue circles) with the respective cell count indicated by brightness, where darker colors indicate lower cell counts.
    Additionally, we include \num{394}~natural samples without known cell counts (gray circles). Note that the circles get darker from left to right, so this direction (determined to be important by UMAP) directly corresponds to the cell count.
    At the same time, natural and synthetic samples are not separated, which would contradict a truly shared representation between the two types of images.}
    \label{fig:umap}
\end{figure}

The ability of \metvae{} to translate between natural and synthetic images already demonstrates that the representation learned by the autoencoder is indeed semantically shared between the two types of images (natural and synthetic). We can go further and visualize the shared representation.
Because each image is encoded as a \num{256}-dimensional vector,
we need to drastically reduce the dimensionality to do so.
\newpage
\emph{Uniform Manifold Approximation and Projection} (UMAP)~\cite{McInnes2018UMAP} computes a topology-preserving embedding and has recently established itself as the state of the art for dimensionality reduction.
In the resulting embedding (see \cref{fig:umap}) we see that synthetic and natural data occupy the same space.
Moreover, by coloring according to cell counts,
we see that an encoding's position is directly related to the number of cells present in the respective image.
Again, this property is shared between natural and synthetic images.

\section{Discussion}
With this architecture we do not only provide a workflow for offline image recognition of suspension cells, it also lays the foundation for automated cell counting of microfluidic cell cultivations with single-cell resolution obtained by live cell imaging and can be extended to continuously estimate cell counts in real-time for experimentation monitoring and prediction of additional relevant information (\eg, survivability of a culture, growth rate or doubling time etc.).
Since the technique works independently of the actual cell size of the analyzed organism, it is possible to generate proxy data for completely different kinds of cells (\eg, bacterial) with slight adjustments to the synthetic data generator.
To reduce the amount of hand crafting meta-parameters further, the Gaussian process regressor can be extended, to allow provision of an easy to use tool, to be operated by biologists as an end-to-end solution for cell counting in live cell imaging procedures.

The novel \metvae{} presented in this paper can be further expanded, \eg, by implementing an interactive manual counting to improve performance similar to \cite{Arteta2014}, which could decrease the training time or even improve the overall results.
As the presumably biggest challenge to overcome, overlapping of cells can be tackled with the methods presented in \cite{Arteta2016}.
Another approach is to extend on data augmentation instead of explicit regularization~\cite{Hernandez2018}.

\subsection{Limitations \& Future Work}
\label{sec:limitations}
For cells that appear very small in the images, the architecture currently can not deliver low error estimations, especially if the cells are only few pixels big in the downscaled version of the images.
For bright-field data, cells sticking to the walls are a bigger hindrance to precise estimations, since, unlike for phase-contrast cells, there often is little to no visual hint for the boundary between the chamber border and the cell membrane.
These problems can presumably be overcome with relative ease by more image preprocessing, like gray-value equalization.

Due to the high number of hyperparameter choices combined with the on average \num{5.6}~seconds per epoch training time, a full training cycle of \num{50000}~epochs takes around three days per parameter set, the Gaussian Process regression is currently infeasible to be used to optimize all hyperparameters. The architecture was trained on a shared cluster that consists of a total of \num{28}~compute nodes, each with two Intel Xeon Silver 4114, \num{404}~GB of RAM and eight NVIDIA Tesla V100 SXM2 32GB.
It has proven useful to parallelize multiple runs and create a new range of hyperparameter options based on the best results of the previous iteration.
The technique shows promise to reduce the number of training cycle iterations needed but currently requires alternating between automation and handcrafting.

\AtNextBibliography{\raggedright\small}
\printbibliography

\end{document}